\documentclass[11pt]{article}
\usepackage{coling2018}
\usepackage{times}
\usepackage{url}
\usepackage{latexsym}
\usepackage{color}
\usepackage{booktabs}
\usepackage{cleveref}
\usepackage{tikz}
\usetikzlibrary{shapes.multipart}
\usetikzlibrary{positioning}
\usepackage{ifthen}
\usepackage{todonotes}

\usepackage[normalem]{ulem}

\title{A Survey on Recent Advances in Named Entity Recognition from Deep Learning models}

\author{Vikas Yadav \\
  University of Arizona \\
  {\tt vikasy@email.arizona.edu} \\\And
  Steven Bethard \\
  University of Arizona \\
  {\tt bethard@email.arizona.edu} \\}

\begin{document}
\maketitle
\begin{abstract}
\vspace{-0.25\baselineskip} 
Named Entity Recognition (NER) is a key component in NLP systems for question answering, information retrieval, relation extraction, etc. NER systems have been studied and developed widely for decades, but accurate systems using deep neural networks (NN) have only been introduced in the last few years. We present a comprehensive survey of deep neural network architectures for NER, and contrast them with previous approaches to NER based on feature engineering and other supervised or semi-supervised learning algorithms. Our results highlight the improvements achieved by neural networks, and show how incorporating some of the lessons learned from past work on feature-based NER systems can yield further improvements.
\end{abstract}

\section{Introduction}
\label{intro}

%
%
\blfootnote{
    %
    %
    \hspace{-0.65cm}  
    This work is licenced under a Creative Commons Attribution 4.0 International Licence. Licence details: \url{http://creativecommons.org/licenses/by/4.0/}
    %
    %
    %
    %
}

Named entity recognition is the task of identifying named entities like person, location, organization, drug, time, clinical procedure, biological protein, etc. in text. NER systems are often used as the first step in question answering, information retrieval, co-reference resolution, topic modeling, etc. Thus it is important to highlight recent advances in named entity recognition, especially recent neural NER architectures which have achieved state of the art performance with minimal feature engineering.

The first NER task was organized by \newcite{grishman1996message} in the Sixth Message Understanding Conference. Since then, there have been numerous NER tasks \cite{tjong2003introduction,sang2002introduction,multilingual2017first,segura2013semeval,bossy2013bionlp,I2B2_license}.
Early NER systems were based on handcrafted rules, lexicons, orthographic features and ontologies.
These systems were followed by NER systems based on feature-engineering and machine learning \cite{nadeau2007survey}.
Starting with \newcite{collobert2011natural}, neural network NER systems with minimal feature engineering have become popular.
Such models are appealing because they typically do not require domain specific resources like lexicons or ontologies, and are thus poised to be more domain independent.
Various neural architectures have been proposed, mostly based on some form of recurrent neural networks (RNN) over characters, sub-words and/or word embeddings. 

We present a comprehensive survey of recent advances in named entity recognition.
We describe knowledge-based and feature-engineered NER systems that combine in-domain knowledge, gazetteers, orthographic and other features with supervised or semi-supervised learning.
We contrast these systems with neural network architectures for NER based on minimal feature engineering, and compare amongst the neural models with different representations of words and sub-word units.
We show in \Cref{table:resultconll} and \Cref{table:DrugNER-official} and discuss in \Cref{Discussion}
how neural NER systems have improved performance over past works including supervised, semi-supervised, and knowledge based NER systems.
For example, NN models on news corpora improved the previous state-of-the-art by 1.59\% in Spanish, 2.34\% in German, 0.36\% in English, and 0.14\%, in Dutch, without any external resources or feature engineering.
We provide resources, including links to shared tasks on NER, and links to the code for each category of NER system. 
To the best of our knowledge, this is the first survey focusing on neural architectures for NER, and comparing to previous feature-based systems.  

We first discuss previous summary research on NER in \cref{prev_studies}. Then we explain our selection criterion and methodology for selecting which systems to review in \cref{method}. We highlight standard, past and recent NER datasets (from shared tasks and other research) in \cref{datasets} and evaluation metrics in \cref{Evaluation}. We then describe NER systems in \cref{categories} categorized into knowledge-based (\cref{Knowledge-based}), bootstrapped (\cref{Bootstrapped}), feature-engineered (\cref{Feature-engineered}) and neural networks (\cref{NN_system}).

\section{Previous surveys}
\label{prev_studies}

The first comprehensive NER survey was \newcite{nadeau2007survey}, which covered a variety of supervised, semi-supervised and unsupervised NER systems, highlighted common features used by NER systems during that time, and explained NER evaluation metrics that are still in use today.
\newcite{MLNERsurvey} presented a more recent NER survey that also included supervised, semi-supervised, and unsupervised NER systems, and included a few introductory neural network NER systems.
There have also been surveys focused on NER systems for specific domains and languages, including biomedical NER, \cite{leaman2008banner}, Chinese clinical NER \cite{Chinese_clinicalNER}, Arabic NER \cite{Arabic2014survey,etaiwi2017statistical}, and NER for Indian languages \cite{Indian_survey_NER}.

The existing surveys primarily cover feature-engineered machine learning models (including supervised, semi-supervised, and unsupervised systems), and mostly focus on a single language or a single domain.
There is not yet, to our knowledge, a comprehensive survey of modern neural network NER systems, nor is there a survey that compares feature engineered and neural network systems in both multi-lingual (CoNLL 2002 and CoNLL 2003) and multi-domain (e.g., news and medical) settings. 

\section{Methodology}
\label{method}


To identify articles for this survey, we searched Google, Google Scholar, and Semantic Scholar.
Our query terms included \textit{named entity recognition}, \textit{neural architectures for named entity recognition}, \textit{neural network based named entity recognition models}, \textit{deep learning models for named entity recognition}, etc.
We sorted the papers returned from each query by citation count and read at least the top three, considering a paper for our survey if it either introduced a neural architecture for named entity recognition, or represented a top-performing model on an NER dataset.
We included an article presenting a neural architecture only if it was the first article to introduce the architecture; otherwise, we traced citations back until we found the original source of the architecture. We followed the same approach for feature-engineering NER systems.
We also included articles that implemented these systems for different languages or domain. 
In total, 154 articles were reviewed and 83 articles were selected for the survey.


\section{NER datasets}
\label{datasets}

Since the first shared task on NER \cite{grishman1996message}\footnote{Shared task: \url{https://www-nlpir.nist.gov/related_projects/muc/}}, many shared tasks and datasets for NER have been created. CoNLL 2002 \cite{sang2002introduction}\footnote{Shared task: \url{https://www.clips.uantwerpen.be/conll2002/ner/}} and CoNLL 2003 \cite{tjong2003introduction}\footnote{Shared task: \url{https://www.clips.uantwerpen.be/conll2003/ner/}} were created from newswire articles in four different languages (Spanish, Dutch, English, and German) and focused on 4 entities - PER (person), LOC (location), ORG (organization) and MISC (miscellaneous including all other types of entities).  

NER shared tasks have also been organized for a variety of other languages, including Indian languages \cite{IndianlangNER}, Arabic \cite{Arabic2014survey}, German \cite{German_NER}, and slavic languages \cite{multilingual2017first}.
The named entity types vary widely by source of dataset and language. For example, \newcite{IndianlangNER}'s southeast Asian language data has named entity types person, designation, temporal expressions, abbreviations, object number, brand, etc.
\newcite{German_NER}'s data, which is based on German wikipedia and online news, has named entity types similar to that of CoNLL 2002 and 2003: PERson, ORGanization, LOCation and OTHer.
The shared task\footnote{Shared task: \url{http://bsnlp.cs.helsinki.fi/shared_task.html}} organized by
\newcite{multilingual2017first} covering 7 slavic languages (Croatian, Czech, Polish, Russian, Slovak, Slovene, Ukrainian) also has person, location, organization and miscellaneous as named entity types. 

In the biomedical domain, \newcite{BioNERkim2004introduction} organized a BioNER task on MedLine abstracts, focusing on protien, DNA, RNA and cell attribute entity types. \newcite{uzuner2007evaluating} presented a clinical note de-identification task that required NER to locate personal patient data phrases to be anonymized.
The 2010 I2B2 NER task\footnote{Shared task: \url{https://www.i2b2.org/NLP/Relations/}} \cite{I2B2_license}, which considered clinical data, focused on clinical problem, test and treatment entity types.
\newcite{segura2013semeval} organized a Drug NER shared task\footnote{Shared task: \url{https://www.cs.york.ac.uk/semeval-2013/task9/index.html}} as part of SemEval 2013 Task 9, which focused on drug, brand, group and drug\_n (unapproved or new drugs) entity types.
\cite{CHEMdnerkrallinger2015} introduced the similar CHEMDNER task 
\footnote{Similar datasets can be found here: \url{http://www.biocreative.org}} focusing more on chemical and drug entities like trivial, systematic, abbreviation, formula, family, identifier, etc.
Biology and microbiology NER datasets\footnote{Shared task: \url{http://2016.bionlp-st.org/tasks/bb2}} \cite{Biologyhirschman2005overview,bossy2013bionlp,BBdeleger2016overview} have been collected from PubMed and biology websites, and focus mostly on bacteria, habitat and geo-location entities. In biomedical NER systems, segmentation of clinical and drug entities is considered to be a difficult task because of complex orthographic structures of named entities \cite{liu2015effects}.

NER tasks have also been organized on social media data, e.g., Twitter, where the performance of classic NER systems degrades due to issues like variability in orthography and presence of grammatically incomplete sentences \cite{Twitter_NER}. Entity types on Twitter are also more variable (person, company, facility, band, sportsteam, movie, TV show, etc.) as they are based on user behavior on Twitter. 


Though most named entity annotations are flat, some datasets include more complex structures. \newcite{ohta2002genia} constructed a dataset of nested named entities, where one named entity can contain another. \newcite{strassel2003multilingual} highlighted both entity and entity head phrases.
And discontinuous entities are common in chemical and clinical NER datasets \cite{CHEMdnerkrallinger2015}.
\newcite{eltyeb2014chemical} presented an survey of various NER systems developed for such NER datasets with a focus on chemical NER.

\section{NER evaluation metrics}
\label{Evaluation}

\newcite{grishman1996message} scored NER performance based on \textit{type}, whether the predicted label was correct regardless of entity boundaries, and \textit{text}, whether the predicted entity boundaries were correct regardless of the  label.
For each score category,
\textit{precision} was defined as the number of entities a system predicted correctly divided by the number that the system predicted, 
recall was defined as the number of entities a system predicted correctly divided by the number that were identified by the human annotators,
and (micro) F-score was defined as the harmonic mean of precision and recall from both type and text. 

The \textit{exact match} metrics introduced by CoNLL \cite{tjong2003introduction,sang2002introduction} considers a prediction to be correct only when the predicted label for the complete entity is matched to exactly the same words as the gold label of that entity. CoNLL also used (micro) F-score, taking the harmonic mean of the exact match precision and recall.

The \textit{relaxed F1} and \textit{strict F1} metrics have been used in many NER shared tasks \cite{segura2013semeval,CHEMdnerkrallinger2015,bossy2013bionlp,BBdeleger2016overview}.
Relaxed F1 considers a prediction to be correct as long as part of the named entity is identified correctly.
Strict F1 requires the character offsets of a prediction and the human annotation to match exactly.
In these data, unlike CoNLL, word offsets are not given, so relaxed F1 is intended to allow comparison despite different systems having different word boundaries due to different segmentation techniques \cite{liu2015effects}.

\section{NER systems}
\label{categories}

\subsection{Knowledge-based systems}
\label{Knowledge-based}
Knowledge-based NER systems do not require annotated training data as they rely on lexicon resources and domain specific knowledge.
These work well when the lexicon is exhaustive, but fail, for example, on every example of the drug\_n class in the DrugNER dataset \cite{segura2013semeval}, since drug\_n is defined as unapproved or new drugs, which are by definition not in the DrugBank dictionaries \cite{Drugbank}.
Precision is generally high for knowledge-based NER systems because of the lexicons, but recall is often low due to domain and language-specific rules and incomplete dictionaries. Another drawback of knowledge based NER systems is the need of domain experts for constructing and maintaining the knowledge resources. 

\subsection{Unsupervised and bootstrapped systems}
\label{Bootstrapped}

Some of the earliest systems required very minimal training data.
\newcite{collins1999unsupervised} used only labeled seeds, and 7 features including orthography (e.g., capitalization), context of the entity, words contained within named entities, etc. for classifying and extracting named entities.
\newcite{etzioni2005unsupervised_WEBner} proposed an unsupervised system to improve the recall of NER systems applying 8 generic pattern extractors to open web text, e.g., \textit{NP “is a” $<$class1$>$}, \textit{NP1 “such as” NPList2}.
\newcite{nadeau2006unsupervised} presented an unsupervised system for gazetteer building and named entity ambiguity resolution based on \newcite{etzioni2005unsupervised_WEBner} and \newcite{collins1999unsupervised} that combined an extracted gazetteer with commonly available gazetteers to achieve F-scores  of 88\%, 61\%, and 59\% on MUC-7 \cite{MUC7} location, person, and organization entities, respectively.

\newcite{zhangBIOmedunsupervised} used shallow syntactic knowledge and inverse document frequency (IDF) for an unsupervised NER system on biology \cite{BioNERkim2004introduction} and medical \cite{I2B2_license} data, achieving 53.8\% and 69.5\% accuracy, respectively. Their model
uses seeds to discover text having potential named entities, detects noun phrases and filters any with low IDF values, and feeds the filtered list to a classifier \cite{alfonseca2002unsupervised} to predict named entity tags.

\subsection{Feature-engineered supervised systems}
\label{Feature-engineered}

Supervised machine learning models learn to make predictions by training on example inputs and their expected outputs, and can be used to replace human curated rules. Hidden Markov Models (HMM), Support Vector Machines (SVM), Conditional Random Fields (CRF), and decision trees were common machine learning systems for NER. 

\newcite{HMMzhou2002named} used HMM \cite{HMM1986introduction,HMM1997nymble} an NER system on MUC-6 and MUC-7 data, achieving 96.6\% and 94.1\% F score, respectively. They included 11 orthographic features (1 numeral, 2 numeral, 4 numeral, all caps, numerals and alphabets, contains underscore or not, etc.) a list of trigger words for the named entities (e.g., 36 trigger words and affixes, like \textit{river}, for the location entity class), and a list of words (10000 for the person entity class) from various gazetteers.

\newcite{CoNLL2002markov} compared the HMM with Maximum Entropy (ME) by adding multiple features. 
Their best model included capitalization, whether a word was the first in a sentence, whether a word had appeared before with a known last name, and 13281 first names collected from various dictionaries.
The model achieved 73.66\%, 68.08\% Fscore on Spanish and Dutch CoNLL 2002 dataset respectively.

The winner of CoNLL 2002 \cite{CoNLL2002Winner} used binary AdaBoost classifiers, a boosting algorithm that combines small fixed-depth decision trees \cite{adaBoost2013explaining}. They used features like capitalization, trigger words, previous tag prediction, bag of words, gazetteers, etc. to represent simple binary relations and these relations were used in conjunction with previously predicted labels. They achieved 81.39\% and 77.05\% F scores on the Spanish and Dutch CoNLL 2002 datasets, respectively.

\newcite{SVM_CoNLL2003} implemented a SVM model on the CoNLL 2003 dataset and CMU seminar documents. They experimented with multiple window sizes, features (orthographic, prefixes suffixes, labels, etc.) from neighboring words, weighting neighboring word features according to their position, and class weights to balance positive and negative class. They used two SVM classifiers, one for detecting named entity starts and one for detecting ends.
They achieved 88.3\% F score on the English CoNLL 2003 data. 

On the MUC6 data, \newcite{SVM2002_MUC6} used part-of-speech (POS) tags, orthographic features, a window of 3 words to the left and to the right of the central word, and tags of the last 3 words as features to the SVM. The final tag was decided by the voting of multiple one-vs-one SVM outputs.

\newcite{SEMIsup_state_art_2005} implemented structural learning \cite{ando2005framework} to divide the main task into many auxiliary tasks, for example, predicting labels by looking just at the context and masking the current word.
The best classifier for each auxiliary task was selected based on its confidence.
This model had achieved 89.31\% and 75.27\% F score on English and German, respectively.  

\newcite{agerri2016robust} developed a semi-supervised system\footnote{Code: \url{https://github.com/ixa-ehu/ixa-pipe-nerc}} by presenting NER classifiers with features including orthography, character n-grams, lexicons
, prefixes, suffixes, bigrams, trigrams, and unsupervised cluster features from the Brown corpus, Clark corpus and k-means clustering of open text using word embeddings \cite{WordEmb_mikolov2013}. They achieved near state of the art performance on CoNLL datasets: 84.16\%, 85.04\%, 91.36\%, 76.42\% on Spanish, Dutch, English, and German, respectively.  

In DrugNER \cite{segura2013semeval}, \newcite{liu2015effects} achieved state-of-the-art results by using a CRF with features like lexicon resources from Food and Drug Administration (FDA), DrugBank, Jochem \cite{hettne2009dictionary} and word embeddings (trained on a MedLine corpus). For the same task, \newcite{rocktaschel2013wbi} used a CRF with features constructed from dictionaries (e.g., Jochem \cite{hettne2009dictionary}), ontologies (ChEBI ontologies), prefixes-suffixes from chemical entities, etc.

\subsection{Feature-inferring neural network systems}
\label{NN_system}
\newcite{collobert2008unified} proposed one of the first neural network architectures for NER, with feature vectors constructed from orthographic features (e.g., capitalization of the first character), dictionaries and lexicons.
Later work replaced these manually constructed feature vectors with word embeddings \cite{collobert2011natural}, which are representations of words in $n$-dimensional space, typically learned over large collections of unlabeled data through an unsupervised process such as the skip-gram model \cite{WordEmb_mikolov2013}.
Studies have shown the importance of such pre-trained word embeddings for neural network based NER systems \cite{Emb_help_2017deep}, and similarly for pre-trained character embeddings in character-based languages like Chinese \cite{li2015component,yin2016multi}. 

Modern neural architectures for NER can be broadly classified into categories depending upon their representation of the words in a sentence. For example, representations may be based on words, characters, other sub-word units or any combination of these.

\subsubsection{Word level architectures}
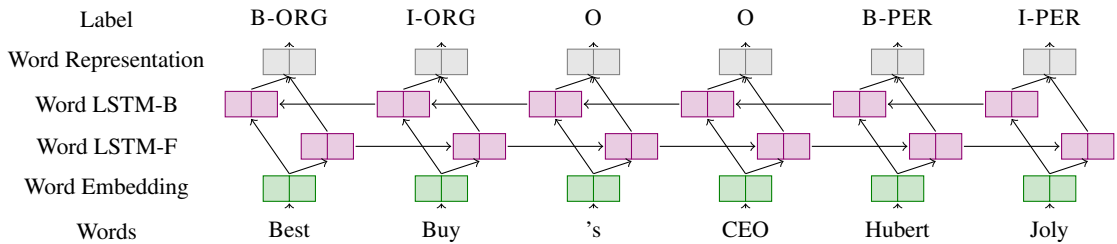
\begin{figure}
\begin{tikzpicture}[
  scale=0.8, every node/.style={scale=0.8},
  hid/.style 2 args={
    rectangle split,
    rectangle split horizontal,
    draw=#2,
    rectangle split parts=#1,
    fill=#2!20,
    outer sep=1pt}]
  \pgfmathsetmacro{\height}{0.7}
  \pgfmathsetmacro{\width}{1.25}
  \node at (-0.5, +1*\height) {Words};
  \node at (-0.5, +2*\height) {Word Embedding};
  \node at (-0.5, +3*\height) {Word LSTM-F};
  \node at (-0.5, +4*\height) {Word LSTM-B};
  \node at (-0.5, +5*\height) {Word Representation};
  \node at (-0.5, +6*\height) {Label};
  \foreach \label/\start/\word [count=\labelIndex from 1] in {B-ORG/2/Best,I-ORG/4/Buy,O/6/'s,O/8/CEO,B-PER/10/Hubert,I-PER/12/Joly} {
    
    \node (fwi\labelIndex) at (\width*\start, +1*\height) {\strut\word};
    \node[hid={2}{green!50!black}] (fw\labelIndex) at (\width*\start, +2*\height) {};
    \draw[->] (fwi\labelIndex.north) -> (fw\labelIndex.south);

    \node[hid={2}{purple!75!blue}] (wrf\labelIndex) at (\width*\start+0.5*\width, +3*\height) {};
    \node[hid={2}{purple!75!blue}] (wrb\labelIndex) at (\width*\start-0.5*\width, +4*\height) {};
    \draw[->] (fw\labelIndex.north) -> (wrf\labelIndex.south);
    \draw[->] (fw\labelIndex.north) -> (wrb\labelIndex.south);
    \ifthenelse{\labelIndex = 1}{}{
      \pgfmathtruncatemacro{\lastConceptIndex}{add(\labelIndex,-1)}
      \draw[->] (wrf\lastConceptIndex.east) -> (wrf\labelIndex.west);
      \draw[<-] (wrb\lastConceptIndex.east) -> (wrb\labelIndex.west);
    }

    \node[hid={2}{gray}] (wc\labelIndex) at (\width*\start, +5*\height) {};
    \draw[->] (wrf\labelIndex.north) -> (wc\labelIndex.south);
    \draw[->] (wrb\labelIndex.north) -> (wc\labelIndex.south);

    \node (o\labelIndex) at (\width*\start, +6*\height) {\strut\sc\label};
    \draw[->] (wc\labelIndex.north) -> (o\labelIndex.south);
  }
\end{tikzpicture}
\vspace*{-0.75\baselineskip} 
\caption{Word level NN architecture for NER}
\label{figure:Word_NN}
\end{figure}

In this architecture, the words of a sentence are given as input to Recurrent Neural Networks (RNN) and each word is represented by its word embedding, as shown in \Cref{figure:Word_NN}. 

The first word-level NN model was proposed by \newcite{collobert2011natural}\footnote{Code: \url{https://ronan.collobert.com/senna/}}. The architecture was similar to the one shown in \Cref{figure:Word_NN}, but a convolution layer was used instead of the Bi-LSTM layer and the output of the convolution layer was given to a CRF layer for the final prediction. The authors achieved 89.59\% F1 score on English CoNLL 2003 dataset by including gazetteers and SENNA embeddings. 

\newcite{WORDhuang2015bidirectional} presented a word LSTM model (\Cref{figure:Word_NN}) and showed that adding a CRF layer to the top of the word LSTM improved performance, achieving 84.26\% F1 score on English CoNLL 2003 dataset.
Similar systems were applied to other domains:
DrugNER by \newcite{chalapathy2016} achieving 85.19\% F1 score (under an unofficial evaluation) on MedLine test data \cite{segura2013semeval},
and medical NER by \newcite{xu2017bidirectional} achieving 80.22\% F1 on disease NER corpus using this architecture.
In similar tasks, \newcite{Goldberg2016multilingual} implemented the same model for multilingual POS tagging. 

With slight variations, \newcite{yan2016multilingual} implemented word level feed forward NN, bi-directional LSTM (bi-LSTM) and window bi-LSTM for NER of English, German and Arabic. They also highlighted the performance improvement after adding various features like CRF, case, POS, word embeddings and achieved 88.91\% F1 score on English and 76.12\% on German.

\subsubsection{Character level architectures}

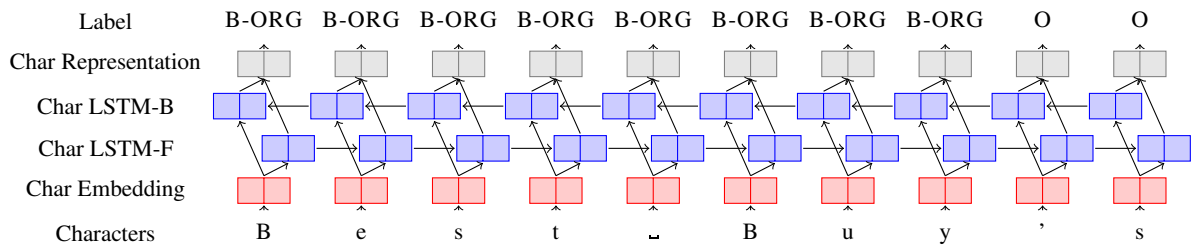
\begin{figure}
\begin{tikzpicture}[
  scale=0.8, every node/.style={scale=0.8},
  hid/.style 2 args={
    rectangle split,
    rectangle split horizontal,
    draw=#2,
    rectangle split parts=#1,
    fill=#2!20,
    outer sep=1pt}]
  \pgfmathsetmacro{\height}{0.7}
  \pgfmathsetmacro{\width}{0.8}
  \node at (-1, +1*\height) {Characters};
  \node at (-1, +2*\height) {Char Embedding};
  \node at (-1, +3*\height) {Char LSTM-F};
  \node at (-1, +4*\height) {Char LSTM-B};
  \node at (-1, +5*\height) {Char Representation};
  \node at (-1, +6*\height) {Label};
  \foreach \label/\start/\word [count=\labelIndex from 1] in {B-ORG/2/B,B-ORG/4/e,B-ORG/6/s,B-ORG/8/t,B-ORG/10/\textvisiblespace,B-ORG/12/B,B-ORG/14/u,B-ORG/16/y,O/18/',O/20/s} {
    
    \node (fci\labelIndex) at (\width*\start, +1*\height) {\strut\word};
    \node[hid={2}{red}] (fc\labelIndex) at (\width*\start, +2*\height) {};
    \draw[->] (fci\labelIndex.north) -> (fc\labelIndex.south);

    \node[hid={2}{blue}] (crf\labelIndex) at (\width*\start+0.5*\width, +3*\height) {};
    \node[hid={2}{blue}] (crb\labelIndex) at (\width*\start-0.5*\width, +4*\height) {};
    \draw[->] (fc\labelIndex.north) -> (crf\labelIndex.south);
    \draw[->] (fc\labelIndex.north) -> (crb\labelIndex.south);
    \ifthenelse{\labelIndex = 1}{}{
      \pgfmathtruncatemacro{\lastConceptIndex}{add(\labelIndex,-1)}
      \draw[->] (crf\lastConceptIndex.east) -> (crf\labelIndex.west);
      \draw[<-] (crb\lastConceptIndex.east) -> (crb\labelIndex.west);
    }

    \node[hid={2}{gray}] (c\labelIndex) at (\width*\start, +5*\height) {};
    \draw[->] (crf\labelIndex.north) -> (c\labelIndex.south);
    \draw[->] (crb\labelIndex.north) -> (c\labelIndex.south);

    \node (o\labelIndex) at (\width*\start, +6*\height) {\strut\sc\label};
    \draw[->] (c\labelIndex.north) -> (o\labelIndex.south);
  }
\end{tikzpicture}
\vspace*{-0.75\baselineskip} 
\caption{Character level NN architecture for NER}
\label{figure:Char_NN}
\end{figure}

In this model, a sentence is taken to be a sequence of characters. This sequence is passed through an RNN, predicting labels for each character (\Cref{figure:Char_NN}).
Character labels transformed into word labels via post processing. 
The potential of character NER neural models was first highlighted by \newcite{NYU2016character} using highway networks over convolution neural networks (CNN) on character sequences of words and then using another layer of LSTM + softmax for the final predictions. 

This model was implemented by \newcite{pham2017end} for Vietnamese NER and achieved 80.23\% F-score on \newcite{nguyen2016vietnamese}'s Vietnamese test data.
Character models were also used in various other languages like Chinese \cite{character_chinese2016} where it has achieved near state of the art performance.

\newcite{Character_multilingual2016} proposed CharNER \footnote{Code: \url{https://github.com/ozanarkancan/char-ner}} which implemented the character RNN model for NER on 7 different languages. In this character model, tag prediction over characters were converted to word tags using Viterbi decoder\cite{viterbi1973} achieving 82.18\% on Spanish, 79.36\% on Dutch, 84.52\% on English and 70.12\% on German CoNLL datasets. They also achieved 78.72 on Arabic, 72.19 on Czech and 91.30 on Turkish.
\newcite{CHAR_POS_SOA2015} proposed word representation using RNN (Bi-LSTM) over characters of the word and achieved state of the art results on POS task using this representation in multiple languages including 97.78\% accuracy on English PTB\cite{PTB_English}.

\newcite{gillick2015multilingual} implemented sequence to sequence model (Byte to Span- BTS) using encoder decoder architecture over sequence of characters of words in a window of 60 characters. Each character was encoded in bytes and BTS achieved high performance on CoNLL 2002 and 2003 dataset without any feature engineering. BTS achieved 82.95\%, 82.84\%,86.50\%,76.22\% Fscore on Spanish, Dutch, English and German CoNLL datasets respectively.

\subsubsection{Character+Word level architectures}

\begin{figure}
\begin{tikzpicture}[
  scale=0.8, every node/.style={scale=0.8},
  hid/.style 2 args={
    rectangle split,
    rectangle split horizontal,
    draw=#2,
    rectangle split parts=#1,
    fill=#2!20,
    outer sep=1pt}]
  \pgfmathsetmacro{\height}{0.7}
  \pgfmathsetmacro{\width}{1.25}
  \node (i0) at (-0.5, -3*\height) {Characters};
  \node (ce0) at (-0.5, -2*\height) {Char Embedding};
  \node (crf0) at (-0.5, -1*\height) {Char LSTM-F};
  \node (crb0) at (-0.5, 0*\height) {Char LSTM-B};
  \node (wf0) at (-0.5, +1*\height) {Word Features};
  \node (wf0) at (-0.5, +2*\height) {Word Representation};
  \node (wr0) at (-0.5, +3*\height) {Word LSTM-F};
  \node (wr0) at (-0.5, +4*\height) {Word LSTM-B};
  \node (wc0) at (-0.5, +5*\height) {Word CRF};
  \node (o0) at (-0.5, +6*\height) {Label};
  \foreach \label/\start/\word/\chars [count=\labelIndex from 1] in {B-ORG/1/Best/{B,e,s,t},I-ORG/6/Buy/{B,u,y},O/11/'s/{',s}} {
    \foreach \char [count=\charIndex from 1] in \chars {
    
      \node (i\labelIndex-\charIndex) at (\width*\start+\width*\charIndex, -3*\height) {\strut\char};
      
      \node[hid={2}{red}] (ce\labelIndex-\charIndex) at (\width*\start+\width*\charIndex, -2*\height) {};    
      \draw[->] (i\labelIndex-\charIndex.north) -> (ce\labelIndex-\charIndex.south);
      
      \node[hid={2}{blue}] (crf\labelIndex-\charIndex) at (\width*\start+\width*\charIndex+0.5*\width, -1*\height) {};
      \node[hid={2}{blue}] (crb\labelIndex-\charIndex) at (\width*\start+\width*\charIndex-0.5*\width, 0*\height) {};
      \draw[->] (ce\labelIndex-\charIndex.north) -> (crf\labelIndex-\charIndex.south);
      \draw[->] (ce\labelIndex-\charIndex.north) -> (crb\labelIndex-\charIndex.south);
      \ifthenelse{\charIndex = 1}{}{
        \pgfmathtruncatemacro{\lastCharIndex}{add(\charIndex,-1)}
        \draw[->] (crf\labelIndex-\lastCharIndex.east) -> (crf\labelIndex-\charIndex.west);
        \draw[<-] (crb\labelIndex-\lastCharIndex.east) -> (crb\labelIndex-\charIndex.west);
      }
    }
    
    \node (fwi\labelIndex) at (-\width*0.7+\width*\start+\width*\charIndex, +1*\height) {\strut\word};
    \node[hid={2}{green!50!black}] (fw\labelIndex) at (-\width*0.7+\width*\start+\width*\charIndex, +2*\height) {};
    \draw[->] (fwi\labelIndex.north) -> (fw\labelIndex.south);

    \node[hid={2}{blue}] (fc\labelIndex) at (-\width*0+\width*\start+\width*\charIndex, +2*\height) {};
    \draw[->] (crf\labelIndex-\charIndex.north) -> (fc\labelIndex.south);
    \draw[->] (crb\labelIndex-1.north) -> (fc\labelIndex.south);

    \node[hid={2}{purple!75!blue}] (wrf\labelIndex) at (\width*\start+\width*\charIndex+0.5*\width, +3*\height) {};
    \node[hid={2}{purple!75!blue}] (wrb\labelIndex) at (\width*\start+\width*\charIndex-0.5*\width, +4*\height) {};
    \draw[->] (fw\labelIndex.north east) -> (wrf\labelIndex.south);
    \draw[->] (fw\labelIndex.north east) -> (wrb\labelIndex.south);
    \ifthenelse{\labelIndex = 1}{}{
      \pgfmathtruncatemacro{\lastConceptIndex}{add(\labelIndex,-1)}
      \draw[->] (wrf\lastConceptIndex.east) -> (wrf\labelIndex.west);
      \draw[<-] (wrb\lastConceptIndex.east) -> (wrb\labelIndex.west);
    }

    \node[hid={2}{gray}] (wc\labelIndex) at (\width*\start+\width*\charIndex, +5*\height) {};
    \draw[->] (wrf\labelIndex.north) -> (wc\labelIndex.south);
    \draw[->] (wrb\labelIndex.north) -> (wc\labelIndex.south);
    \ifthenelse{\labelIndex = 1}{}{
      \pgfmathtruncatemacro{\lastConceptIndex}{add(\labelIndex,-1)}
      \draw[<->] (wc\lastConceptIndex.east) -> (wc\labelIndex.west);
    }

    \node (o\labelIndex) at (\width*\start+\width*\charIndex, +6*\height) {\strut\sc\label};
    \draw[->] (wc\labelIndex.north) -> (o\labelIndex.south);
  }
\end{tikzpicture}
\vspace*{-0.75\baselineskip} 
\caption{Word+character level NN architecture for NER}
\label{figure:WordChar_NN}
\end{figure}
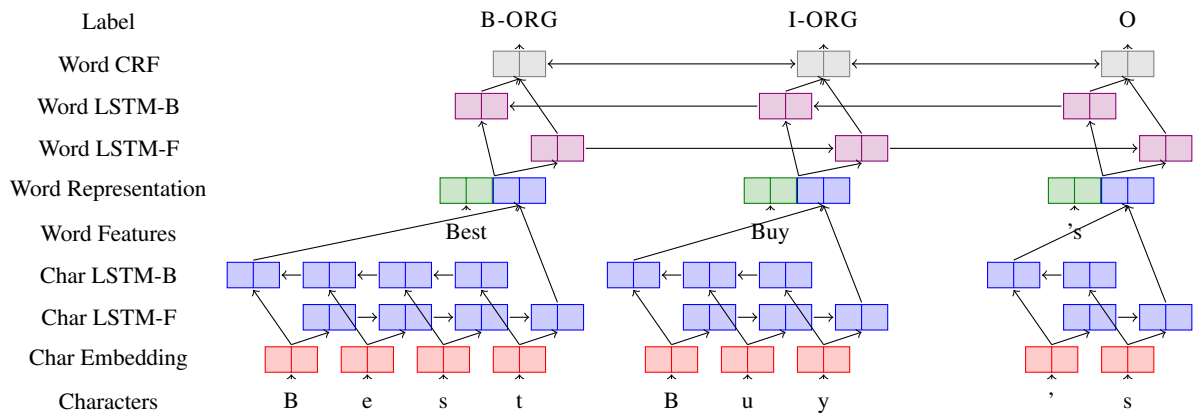

Systems combining word context and the characters of a word have proved to be strong NER systems that need little domain specific knowledge or resources. There are two base models in this category. The \textbf{first type of model} represents words as a combination of a word embedding and a convolution over the characters of the word, follows this with a Bi-LSTM layer over the word representations of a sentence, and finally uses a softmax or CRF layer over the Bi-LSTM to generate labels. The architecture diagram for this model is same as \Cref{figure:WordChar_NN} but with the character Bi-LSTM  replaced with a CNN\footnote{Code: \url{https://github.com/LopezGG/NN_NER_tensorFlow}}. 

\newcite{ma2016end} implemented this model to achieve 91.21\% F1 score on the CoNLL 2003 English dataset and 97.55\% POS-tagging accuracy on the WSJ portion of PTB \cite{PTB_English}. They also showed lower performance by this model for out of vocabulary words.

\newcite{chiu2015named} achieved 91.62\% F1 score on the CoNLL 2003 English dataset and 86.28\% F score on Onto notes 5.0 dataset \cite{ONTO_notes} by adding lexicons and capitalization features
to this model. Lexicon feature were encoded in the form or B(begin), I(inside) or E(end) PER, LOC, ORG and MISC depending upon the match from the dictionary.

This model has also been utilized for NER in languages like Japanese where \newcite{misawa2017character} showed that this architecture outperformed other neural architectures on the \textit{organization} entity class. 

\newcite{Character_twitter_NER} implemented a character+word level NER model for Twitter NER \cite{Twitter_NER} by concatenating a CNN over characters, a CNN over orthographic features of characters, a word embedding, and a word orthographic feature embedding. This concatenated representation is passed through another Bi-LSTM layer and the output is given to CRF for predicting. This model achieved 65.89\% F score on segmentation alone and 52.41\% F score on segmentation and categorization.

\newcite{santos2015boosting} implemented a model with a CNN over the characters of word, concatenated with word embeddings of the central word and its neighbors, fed to a feed forward network, and followed by the Viterbi algorithm to predict labels for each word. The model achieved 82.21\% F score on Spanish CoNLL 2002 data and 71.23\% F score on Portuguese NER data \cite{santos2007Portugese}. 

The \textbf{second type of model} concatenates word embeddings with LSTMs (sometimes bi-directional) over the characters of a word,
passing this representation through another sentence-level Bi-LSTM, and predicting the final tags using a final softmax or CRF layer (\Cref{figure:WordChar_NN}). \newcite{lample2016neural}\footnote{Code: \url{https://github.com/glample/tagger}} introduced this architecture and achieved 85.75\%, 81.74\%, 90.94\%, 78.76\% Fscores on Spanish, Dutch, English and German NER dataset respectively from CoNLL 2002 and 2003. 

\newcite{NeuroNER_MIT} implemented this model in the NeuroNER toolkit\footnote{Code: \url{http://neuroner.com}} with the main goal of providing easy usability and allowing easy plotting of real time performance and learning statistics of the model.
The BRAT annotation tool\footnote{Code: \url{http://brat.nlplab.org/}} is also integrated with NeuroNER to ease the development of NN NER models in new domains. 
NeuroNER achieved 90.50\% F score on the English CoNLL 2003 data.  

\newcite{Emb_help_2017deep} implemented the model for various biomedical NER tasks and achieved higher performance than the majority of other participants. For example, they achieved 83.71 F-score on the CHEMDNER data \cite{CHEMdnerkrallinger2015}.

\newcite{Bharadwaj2016PhonologicallyAN}\footnote{Code: \url{https://github.com/dmort27/epitran}} utilized phonemes (from Epitran) for NER in addition to characters and words. They also utilize attention knowledge over sequence of characters in word which is concatenated with the word embedding and character representation of word.  This model achieved state of the art performance (85.81\% F score) on Spanish CoNLL 2002 dataset.

A slightly improved system focusing on multi-task and multi-lingual joint learning was proposed by \newcite{yang2016multi} where word representation given by GRU (Gated Recurrent Unit) cell over characters plus word embedding was passed through another RNN layer and the output was given to CRF models trained for different tasks like POS, chunking and NER. \newcite{yang2017transfer} further proposed transfer learning for multi-task and multi-learning, and showed small improvements on CoNLL 2002 and 2003 NER data, achieving 85.77\%, 85.19\%, 91.26\% F scores on Spanish, Dutch and English, respectively.

\subsubsection{Character + Word + affix model}

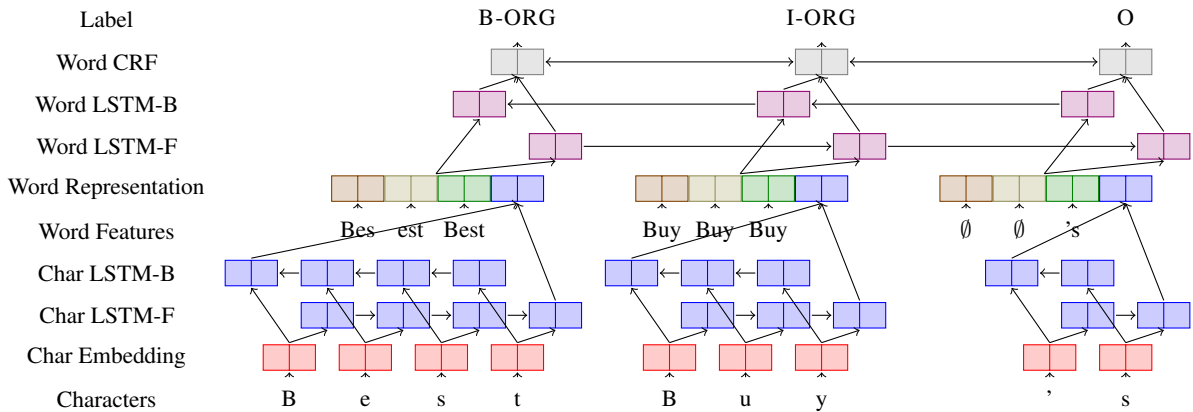
\begin{figure}
\begin{tikzpicture}[
  scale=0.8, every node/.style={scale=0.8},
  hid/.style 2 args={
    rectangle split,
    rectangle split horizontal,
    draw=#2,
    rectangle split parts=#1,
    fill=#2!20,
    outer sep=1pt}]
  \pgfmathsetmacro{\height}{0.7}
  \pgfmathsetmacro{\width}{1.25}
  \node (i0) at (-0.5, -3*\height) {Characters};
  \node (ce0) at (-0.5, -2*\height) {Char Embedding};
  \node (crf0) at (-0.5, -1*\height) {Char LSTM-F};
  \node (crb0) at (-0.5, 0*\height) {Char LSTM-B};
  \node (wf0) at (-0.5, +1*\height) {Word Features};
  \node (wf0) at (-0.5, +2*\height) {Word Representation};
  \node (wr0) at (-0.5, +3*\height) {Word LSTM-F};
  \node (wr0) at (-0.5, +4*\height) {Word LSTM-B};
  \node (wc0) at (-0.5, +5*\height) {Word CRF};
  \node (o0) at (-0.5, +6*\height) {Label};
  \foreach \label/\start/\word/\chars/\prefix/\suffix [count=\labelIndex from 1] in {B-ORG/1/Best/{B,e,s,t}/Bes/est,I-ORG/6/Buy/{B,u,y}/Buy/Buy,O/11/'s/{',s}/$\emptyset$/$\emptyset$} {
    \foreach \char [count=\charIndex from 1] in \chars {
    
      \node (i\labelIndex-\charIndex) at (\width*\start+\width*\charIndex, -3*\height) {\strut\char};
      
      \node[hid={2}{red}] (ce\labelIndex-\charIndex) at (\width*\start+\width*\charIndex, -2*\height) {};    
      \draw[->] (i\labelIndex-\charIndex.north) -> (ce\labelIndex-\charIndex.south);
      
      \node[hid={2}{blue}] (crf\labelIndex-\charIndex) at (\width*\start+\width*\charIndex+0.5*\width, -1*\height) {};
      \node[hid={2}{blue}] (crb\labelIndex-\charIndex) at (\width*\start+\width*\charIndex-0.5*\width, 0*\height) {};
      \draw[->] (ce\labelIndex-\charIndex.north) -> (crf\labelIndex-\charIndex.south);
      \draw[->] (ce\labelIndex-\charIndex.north) -> (crb\labelIndex-\charIndex.south);
      \ifthenelse{\charIndex = 1}{}{
        \pgfmathtruncatemacro{\lastCharIndex}{add(\charIndex,-1)}
        \draw[->] (crf\labelIndex-\lastCharIndex.east) -> (crf\labelIndex-\charIndex.west);
        \draw[<-] (crb\labelIndex-\lastCharIndex.east) -> (crb\labelIndex-\charIndex.west);
      }
    }
    
    \node (fpi\labelIndex) at (-\width*2.1+\width*\start+\width*\charIndex, +1*\height) {\strut\prefix};
    \node[hid={2}{orange!50!black}] (fp\labelIndex) at (-\width*2.1+\width*\start+\width*\charIndex, +2*\height) {};
    \draw[->] (fpi\labelIndex.north) -> (fp\labelIndex.south);
    
    \node (fsi\labelIndex) at (-\width*1.4+\width*\start+\width*\charIndex, +1*\height) {\strut\suffix};
    \node[hid={2}{yellow!50!black}] (fs\labelIndex) at (-\width*1.4+\width*\start+\width*\charIndex, +2*\height) {};
    \draw[->] (fsi\labelIndex.north) -> (fs\labelIndex.south);
    
    \node (fwi\labelIndex) at (-\width*0.7+\width*\start+\width*\charIndex, +1*\height) {\strut\word};
    \node[hid={2}{green!50!black}] (fw\labelIndex) at (-\width*0.7+\width*\start+\width*\charIndex, +2*\height) {};
    \draw[->] (fwi\labelIndex.north) -> (fw\labelIndex.south);

    \node[hid={2}{blue}] (fc\labelIndex) at (-\width*0+\width*\start+\width*\charIndex, +2*\height) {};
    \draw[->] (crf\labelIndex-\charIndex.north) -> (fc\labelIndex.south);
    \draw[->] (crb\labelIndex-1.north) -> (fc\labelIndex.south);

    \node[hid={2}{purple!75!blue}] (wrf\labelIndex) at (\width*\start+\width*\charIndex+0.5*\width, +3*\height) {};
    \node[hid={2}{purple!75!blue}] (wrb\labelIndex) at (\width*\start+\width*\charIndex-0.5*\width, +4*\height) {};
    \draw[->] (fw\labelIndex.north west) -> (wrf\labelIndex.south);
    \draw[->] (fw\labelIndex.north west) -> (wrb\labelIndex.south);
    \ifthenelse{\labelIndex = 1}{}{
      \pgfmathtruncatemacro{\lastConceptIndex}{add(\labelIndex,-1)}
      \draw[->] (wrf\lastConceptIndex.east) -> (wrf\labelIndex.west);
      \draw[<-] (wrb\lastConceptIndex.east) -> (wrb\labelIndex.west);
    }

    \node[hid={2}{gray}] (wc\labelIndex) at (\width*\start+\width*\charIndex, +5*\height) {};
    \draw[->] (wrf\labelIndex.north) -> (wc\labelIndex.south);
    \draw[->] (wrb\labelIndex.north) -> (wc\labelIndex.south);
    \ifthenelse{\labelIndex = 1}{}{
      \pgfmathtruncatemacro{\lastConceptIndex}{add(\labelIndex,-1)}
      \draw[<->] (wc\lastConceptIndex.east) -> (wc\labelIndex.west);
    }

    \node (o\labelIndex) at (\width*\start+\width*\charIndex, +6*\height) {\strut\sc\label};
    \draw[->] (wc\labelIndex.north) -> (o\labelIndex.south);
  }
\end{tikzpicture}
\vspace*{-0.75\baselineskip} 
\caption{Word+character+affix level NN architecture for NER}
\label{figure:OurModel_NN}
\end{figure}


\newcite{yadav2018deep} implemented a model that augments the character+word NN architecture with one of the most successful features from feature-engineering approaches: affixes.
Affix features were used in early NER systems for CoNLL 2002~\cite{sang2002introduction,cucerzan2002language} and 2003~\cite{tjong2003introduction} and for biomedical NER~\cite{SAHA2009905}, but had not been used in neural NER systems.
They extended the \newcite{lample2016neural} character+word model to learn affix embeddings\footnote{Code: \url{https://github.com/vikas95/Pref_Suff_Span_NN}} alongside the word embeddings and character RNNs (\Cref{figure:OurModel_NN}).
They considered all n-gram prefixes and suffixes of words in the training corpus, and selected only those whose frequency was above a threshold, $T$.
Their word+character+affix model achieved 87.26\%, 87.54\%, 90.86\%, 79.01\% on Spanish, Dutch, English and German CoNLL datasets respectively.
\newcite{yadav2018deep} also showed that  affix embeddings capture complementary information to that captured by RNNs over the characters of a word, that selecting only high frequency (realistic) affixes was important, and that embedding affixes was better than simply expanding the other embeddings to reach a similar number of hyper-parameters.

\section{Discussion}
\label{Discussion}

\Cref{table:resultconll} shows the results of all the different categories of systems discussed in \cref{categories} on the CoNLL 2002 and 2003 datasets.
The table also indicates, for each model, whether it makes use of external knowledge like a dictionary or gazetteer.
\Cref{table:DrugNER-official} presents a similar analysis on the DrugNER dataset from SemEval 2013 task 9 \cite{segura2013semeval}.

\begin{table}
\small
\centering
\begin{tabular}{l c c c c c}
\toprule
 {\bf Feature-engineered machine learning systems} & Dict & SP & DU & EN & GE  \\
 \midrule
 \newcite{CoNLL2002Winner} binary AdaBoost classifiers & Yes & 81.39 & 77.05 & - & - \\
 \newcite{CoNLL2002markov} - Maximum Entropy (ME) + features & Yes & 73.66 & 68.08 & - & - \\
 \newcite{SVM_CoNLL2003} SVM with class weights & Yes & - & - & 88.3 & - \\
 \newcite{McLLum_CRF_2014lexicon} CRF & Yes & - & - & 90.90 & - \\
 \newcite{SEMIsup_state_art_2005} Semi-supervised state of the art & No & - & - & 89.31 & 75.27 \\
 \newcite{agerri2016robust} & Yes & \bf{84.16} & \bf{85.04} & \bf{91.36} & \bf{76.42}  \\
 \midrule
  {\bf Feature-inferring neural network word models} \\
 \midrule
 \newcite{collobert2011natural} Vanilla NN +SLL / Conv-CRF & No & - & - & 81.47 & - \\
 \newcite{WORDhuang2015bidirectional} Bi-LSTM+CRF & No & - & - & 84.26 & -\\
 \newcite{yan2016multilingual} Win-BiLSTM (English), FF (German) (Many fets)  & Yes & - & - & 88.91 & \bf{76.12} \\
  \newcite{collobert2011natural} Conv-CRF (SENNA+Gazetteer) & Yes & - & - & 89.59 & - \\
  \newcite{WORDhuang2015bidirectional} Bi-LSTM+CRF+ (SENNA+Gazetteer) & Yes & - & - & \bf{90.10} & -\\

 \midrule
 {\bf Feature-inferring neural network character models}  \\
 \midrule
  \newcite{gillick2015multilingual} -- BTS & No & \bf{82.95} &  \bf{82.84} & \bf{86.50} & \bf{76.22} \\
  \newcite{Character_multilingual2016} CharNER & No & 82.18 & 79.36 & 84.52 & 70.12 \\
  
 \midrule
 {\bf Feature-inferring neural network word + character models} \\
 \midrule
 \newcite{yang2017transfer} & Yes & 85.77 & \bf{85.19} & 91.26 & - \\
 \newcite{Luo_microsoft} & Yes & - & - & 91.20 & - \\
 \newcite{chiu2015named} & Yes & - & - & {\bf 91.62} & - \\
 \newcite{ma2016end} & No & - & - & 91.21 & - \\
 \newcite{santos2015boosting} & No & 82.21 & - & - & -\\
 \newcite{lample2016neural} & No & 85.75& 81.74 &  90.94 & \bf{78.76} \\
 \newcite{Bharadwaj2016PhonologicallyAN} & Yes & \bf{85.81} & - & - & - \\
 \newcite{NeuroNER_MIT} & No & - & - & 90.5 & - \\
 \midrule
 {\bf Feature-inferring neural network word + character + affix models} \\
  \midrule
 Re-implementation of \newcite{lample2016neural} (100 Epochs) & No & 85.34 & 85.27 & 90.24 & 78.44 \\
 \newcite{yadav2018deep}(100 Epochs) & No & 86.92 & 87.50 & 90.69 & 78.56\\
\newcite{yadav2018deep} (150 Epochs) & No & {\bf 87.26} & {\bf 87.54} & 90.86 & {\bf 79.01}\\
\bottomrule

\end{tabular}
\caption{\label{table:resultconll} Comparison of NER systemsin four languages: CoNLL 2002 Spanish (SP), CoNLL 2002 Dutch (DU), CoNLL 2003 English (EN), and CoNLL 2003 German (GE). Dict indicates whether or not the approach makes use of dictionary lookups. Best performance in each category is highlighted in bold.
} 
\end{table}

\begin{table*}
\small
\setlength{\tabcolsep}{0.4em}
\centering
\begin{tabular}{l|c|rrr|rrr|rrr}
\toprule
 &  &\multicolumn{3} {c|}{MedLine (80.10\% )} & \multicolumn{3} {c|}{DrugBank (19.90\% )} & \multicolumn{3} {c}{Complete dataset} \\
\cline{3-5} \cline{6-8} \cline{9-11}
& Dict & P & R & F1 & P & R & F1 & P & R & F1 \\
\midrule
\bf{Feature-engineered machine learning systems}\\
\newcite{rocktaschel2013wbi} & Yes & 60.70 & 55.80 & 58.10 & 88.10 & 87.50 & 87.80 & 73.40 & 69.80 & 71.50 \\

\newcite{liu2015effects} (baseline) & No & - & - & - & - & - & - & 78.41 & 67.78 & 72.71\\

\newcite{liu2015effects} (MED. emb.) & No & - & - & - & - & - & - & 82.70 & 69.68 & 75.63 \\

\newcite{liu2015effects} (state of the art) & Yes & 78.77 & 60.21 & 68.25 & 90.60 & 88.82 & \textbf{89.70} & 84.75 & 72.89 & \textbf{78.37} \\

\midrule
 {\bf NN word model} \\
   
\newcite{chalapathy2016} (relaxed performance) & No & 52.93 & 52.57 & 52.75 & 87.07 & 83.39 & 85.19 & - & - & - \\
\midrule

 {\bf NN word + character model} \\
    
\newcite{yadav2018deep} & No & 73\phantom{.00} & 62\phantom{.00} & 67\phantom{.00} & 87\phantom{.00} & 86\phantom{.00} & 87\phantom{.00} & 79\phantom{.00} & 72\phantom{.00} & 75\phantom{.00} \\
\midrule
 {\bf NN word + character + affix model} \\
    
\newcite{yadav2018deep} & No & 74\phantom{.00} & 64\phantom{.00} & \textbf{69\phantom{.00}} & 89\phantom{.00} & 86\phantom{.00} & 87\phantom{.00} & 81\phantom{.00} & 74\phantom{.00} & 77\phantom{.00} \\

\bottomrule
\end{tabular}
\caption{\label{table:DrugNER-official} DrugNER results on the MedLine and DrugBank test data (80.10\% and 19.90\% of the  test data, respectively). The \newcite{yadav2018deep} experiments report no decimal places because they were run after the end of shared task, and the official evaluation script outputs no decimal places.
} 
\end{table*}

Our first finding from the survey is that feature-inferring NN systems outperform feature-engineered systems, despite the latter's access to domain specific rules, knowledge, features, and lexicons.
For example, the best feature-engineered system for Spanish, \newcite{agerri2016robust}, is 1.59\% below the best feature-inferring neural network system, \cite{lample2016neural}, and 1.65\% below the best neural network system that incorporates lexical resources \cite{Bharadwaj2016PhonologicallyAN}.
Similarly, the best feature-engineered system for German, \newcite{agerri2016robust}, is 2.34\% below the best feature-inferring neural network system, \newcite{lample2016neural}.
The differences are smaller for Dutch and English, but in neither case is the best feature-engineered model better than the best neural network model. In DrugNER, the word+character NN model outperforms the feature engineered system by 8.90\% on MedLine test data and 3.50\% on the overall dataset.

Our next finding is that word+character hybrid models are generally better than both word-based and character-based models.
For example, the best hybrid NN model for English, \newcite{chiu2015named}, is 0.52\% better than the best word-based model, \newcite{WORDhuang2015bidirectional}, and 5.12\% better than the best character-based model, \cite{Character_multilingual2016}.
Similarly, the best hybrid NN model for German, \newcite{lample2016neural}, is 2.64\% better than the best word-based model, \newcite{yan2016multilingual}, and 2.54\% better than the best character-based model, \cite{Character_multilingual2016}. In DrugNER, the word+character hybrid model is better than the word model by 14.25\% on MedLine test data and 1.81\% on DrugBank test data.

Our final finding is that there is still interesting progress to be made by incorporating key features of past feature-engineered models into modern NN architectures.
\newcite{yadav2018deep}'s simple extension of \newcite{lample2016neural} to incorporate affix features yields a very strong new model, achieving a new state-of-the-art in Spanish, Dutch, and German, and performing within 1\% of the best model for English.

\section{Conclusion}
\label{Conclusion}
Our survey of models for named entity recognition, covering both classic feature-engineered machine learning models, and modern feature-inferring neural network models has yielded several important insights.
Neural network models generally outperform feature-engineered models, character+word hybrid neural networks generally outperform other representational choices, and further improvements are available by applying past insights to current neural network models, as shown by the state-of-the-art performance of our proposed affix-based extension of character+word hybrid models.

\bibliographystyle{acl}
\bibliography{main}

\begin{thebibliography}{}

\bibitem[\protect\citename{Agerri and Rigau}2016]{agerri2016robust}
Rodrigo Agerri and German Rigau.
\newblock 2016.
\newblock Robust multilingual named entity recognition with shallow
  semi-supervised features.
\newblock {\em Artificial Intelligence}, 238:63--82.

\bibitem[\protect\citename{Alfonseca and
  Manandhar}2002]{alfonseca2002unsupervised}
Enrique Alfonseca and Suresh Manandhar.
\newblock 2002.
\newblock An unsupervised method for general named entity recognition and
  automated concept discovery.
\newblock In {\em Proceedings of the 1st international conference on general
  WordNet, Mysore, India}, pages 34--43.

\bibitem[\protect\citename{Ando and Zhang}2005a]{SEMIsup_state_art_2005}
Rie~Kubota Ando and Tong Zhang.
\newblock 2005a.
\newblock A framework for learning predictive structures from multiple tasks
  and unlabeled data.
\newblock {\em Journal of Machine Learning Research}, 6(Nov):1817--1853.

\bibitem[\protect\citename{Ando and Zhang}2005b]{ando2005framework}
Rie~Kubota Ando and Tong Zhang.
\newblock 2005b.
\newblock A framework for learning predictive structures from multiple tasks
  and unlabeled data.
\newblock {\em Journal of Machine Learning Research}, 6(Nov):1817--1853.

\bibitem[\protect\citename{Baldwin \bgroup et al.\egroup }2015]{Twitter_NER}
Timothy Baldwin, Marie-Catherine de~Marneffe, Bo~Han, Young-Bum Kim, Alan
  Ritter, and Wei Xu.
\newblock 2015.
\newblock Shared tasks of the 2015 workshop on noisy user-generated text:
  Twitter lexical normalization and named entity recognition.
\newblock In {\em Proceedings of the Workshop on Noisy User-generated Text},
  pages 126--135.

\bibitem[\protect\citename{Benikova \bgroup et al.\egroup }2014]{German_NER}
Darina Benikova, Chris Biemann, and Marc Reznicek.
\newblock 2014.
\newblock Nosta-d named entity annotation for german: Guidelines and dataset.
\newblock In {\em LREC}, pages 2524--2531.

\bibitem[\protect\citename{Bharadwaj \bgroup et al.\egroup
  }2016]{Bharadwaj2016PhonologicallyAN}
Akash Bharadwaj, David~R. Mortensen, Chris Dyer, and Carlos de~Juan~Carbonell.
\newblock 2016.
\newblock Phonologically aware neural model for named entity recognition in low
  resource transfer settings.
\newblock In {\em EMNLP}.

\bibitem[\protect\citename{Bikel \bgroup et al.\egroup }1997]{HMM1997nymble}
Daniel~M Bikel, Scott Miller, Richard Schwartz, and Ralph Weischedel.
\newblock 1997.
\newblock Nymble: a high-performance learning name-finder.
\newblock In {\em Proceedings of the fifth conference on Applied natural
  language processing}, pages 194--201. Association for Computational
  Linguistics.

\bibitem[\protect\citename{Bossy \bgroup et al.\egroup }2013]{bossy2013bionlp}
Robert Bossy, Wiktoria Golik, Zorana Ratkovic, Philippe Bessi{\`e}res, and
  Claire N{\'e}dellec.
\newblock 2013.
\newblock Bionlp shared task 2013--an overview of the bacteria biotope task.
\newblock In {\em Proceedings of the BioNLP Shared Task 2013 Workshop}, pages
  161--169.

\bibitem[\protect\citename{Carreras \bgroup et al.\egroup
  }2002]{CoNLL2002Winner}
Xavier Carreras, Llu{\'\i}s M{\`a}rquez, and Llu{\'\i}s Padr{\'o}.
\newblock 2002.
\newblock Named entity extraction using adaboost, proceedings of the 6th
  conference on natural language learning.
\newblock {\em August}, 31:1--4.

\bibitem[\protect\citename{Chalapathy \bgroup et al.\egroup
  }2016]{chalapathy2016}
Raghavendra Chalapathy, Ehsan~Zare Borzeshi, and Massimo Piccardi.
\newblock 2016.
\newblock An investigation of recurrent neural architectures for drug name
  recognition.
\newblock {\em arXiv preprint arXiv:1609.07585}.

\bibitem[\protect\citename{Chinchor and Robinson}1997]{MUC7}
Nancy Chinchor and Patricia Robinson.
\newblock 1997.
\newblock Muc-7 named entity task definition.
\newblock In {\em Proceedings of the 7th Conference on Message Understanding},
  volume~29.

\bibitem[\protect\citename{Chiu and Nichols}2015]{chiu2015named}
Jason~PC Chiu and Eric Nichols.
\newblock 2015.
\newblock Named entity recognition with bidirectional lstm-cnns.
\newblock {\em arXiv preprint arXiv:1511.08308}.

\bibitem[\protect\citename{Collins and Singer}1999]{collins1999unsupervised}
Michael Collins and Yoram Singer.
\newblock 1999.
\newblock Unsupervised models for named entity classification.
\newblock In {\em 1999 Joint SIGDAT Conference on Empirical Methods in Natural
  Language Processing and Very Large Corpora}.

\bibitem[\protect\citename{Collobert and Weston}2008]{collobert2008unified}
Ronan Collobert and Jason Weston.
\newblock 2008.
\newblock A unified architecture for natural language processing: Deep neural
  networks with multitask learning.
\newblock In {\em Proceedings of the 25th international conference on Machine
  learning}, pages 160--167. ACM.

\bibitem[\protect\citename{Collobert \bgroup et al.\egroup
  }2011]{collobert2011natural}
Ronan Collobert, Jason Weston, L{\'e}on Bottou, Michael Karlen, Koray
  Kavukcuoglu, and Pavel Kuksa.
\newblock 2011.
\newblock Natural language processing (almost) from scratch.
\newblock {\em Journal of Machine Learning Research}, 12(Aug):2493--2537.

\bibitem[\protect\citename{Cucerzan and Yarowsky}2002]{cucerzan2002language}
Silviu Cucerzan and David Yarowsky.
\newblock 2002.
\newblock Language independent ner using a unified model of internal and
  contextual evidence.
\newblock In {\em proceedings of the 6th conference on Natural language
  learning-Volume 20}, pages 1--4. Association for Computational Linguistics.

\bibitem[\protect\citename{Del{\.e}ger \bgroup et al.\egroup
  }2016]{BBdeleger2016overview}
Louise Del{\.e}ger, Robert Bossy, Estelle Chaix, Mouhamadou Ba, Arnaud
  Ferr{\.e}, Philippe Bessieres, and Claire N{\.e}dellec.
\newblock 2016.
\newblock Overview of the bacteria biotope task at bionlp shared task 2016.
\newblock In {\em Proceedings of the 4th BioNLP Shared Task Workshop}, pages
  12--22.

\bibitem[\protect\citename{Dernoncourt \bgroup et al.\egroup
  }2017]{NeuroNER_MIT}
Franck Dernoncourt, Ji~Young Lee, and Peter Szolovits.
\newblock 2017.
\newblock Neuroner: an easy-to-use program for named-entity recognition based
  on neural networks.
\newblock {\em arXiv preprint arXiv:1705.05487}.

\bibitem[\protect\citename{Dong \bgroup et al.\egroup
  }2016]{character_chinese2016}
Chuanhai Dong, Jiajun Zhang, Chengqing Zong, Masanori Hattori, and Hui Di.
\newblock 2016.
\newblock Character-based lstm-crf with radical-level features for chinese
  named entity recognition.
\newblock In {\em Natural Language Understanding and Intelligent Applications},
  pages 239--250. Springer.

\bibitem[\protect\citename{Eltyeb and Salim}2014]{eltyeb2014chemical}
Safaa Eltyeb and Naomie Salim.
\newblock 2014.
\newblock Chemical named entities recognition: a review on approaches and
  applications.
\newblock {\em Journal of cheminformatics}, 6(1):17.

\bibitem[\protect\citename{Etaiwi \bgroup et al.\egroup
  }2017]{etaiwi2017statistical}
Wael Etaiwi, Arafat Awajan, and Dima Suleiman.
\newblock 2017.
\newblock Statistical arabic name entity recognition approaches: A survey.
\newblock {\em Procedia Computer Science}, 113:57--64.

\bibitem[\protect\citename{Etzioni \bgroup et al.\egroup
  }2005]{etzioni2005unsupervised_WEBner}
Oren Etzioni, Michael Cafarella, Doug Downey, Ana-Maria Popescu, Tal Shaked,
  Stephen Soderland, Daniel~S Weld, and Alexander Yates.
\newblock 2005.
\newblock Unsupervised named-entity extraction from the web: An experimental
  study.
\newblock {\em Artificial intelligence}, 165(1):91--134.

\bibitem[\protect\citename{Forney}1973]{viterbi1973}
G~David Forney.
\newblock 1973.
\newblock The viterbi algorithm.
\newblock {\em Proceedings of the IEEE}, 61(3):268--278.

\bibitem[\protect\citename{Gillick \bgroup et al.\egroup
  }2015]{gillick2015multilingual}
Dan Gillick, Cliff Brunk, Oriol Vinyals, and Amarnag Subramanya.
\newblock 2015.
\newblock Multilingual language processing from bytes.
\newblock {\em arXiv preprint arXiv:1512.00103}.

\bibitem[\protect\citename{Grishman and Sundheim}1996]{grishman1996message}
Ralph Grishman and Beth Sundheim.
\newblock 1996.
\newblock Message understanding conference-6: A brief history.
\newblock In {\em COLING 1996 Volume 1: The 16th International Conference on
  Computational Linguistics}, volume~1.

\bibitem[\protect\citename{Habibi \bgroup et al.\egroup
  }2017]{Emb_help_2017deep}
Maryam Habibi, Leon Weber, Mariana Neves, David~Luis Wiegandt, and Ulf Leser.
\newblock 2017.
\newblock Deep learning with word embeddings improves biomedical named entity
  recognition.
\newblock {\em Bioinformatics}, 33(14):i37--i48.

\bibitem[\protect\citename{Hettne \bgroup et al.\egroup
  }2009]{hettne2009dictionary}
Kristina~M Hettne, Rob~H Stierum, Martijn~J Schuemie, Peter~JM Hendriksen,
  Bob~JA Schijvenaars, Erik M~van Mulligen, Jos Kleinjans, and Jan~A Kors.
\newblock 2009.
\newblock A dictionary to identify small molecules and drugs in free text.
\newblock {\em Bioinformatics}, 25(22):2983--2991.

\bibitem[\protect\citename{Hirschman \bgroup et al.\egroup
  }2005]{Biologyhirschman2005overview}
Lynette Hirschman, Alexander Yeh, Christian Blaschke, and Alfonso Valencia.
\newblock 2005.
\newblock Overview of biocreative: critical assessment of information
  extraction for biology.

\bibitem[\protect\citename{Huang \bgroup et al.\egroup
  }2015]{WORDhuang2015bidirectional}
Zhiheng Huang, Wei Xu, and Kai Yu.
\newblock 2015.
\newblock Bidirectional lstm-crf models for sequence tagging.
\newblock {\em arXiv preprint arXiv:1508.01991}.

\bibitem[\protect\citename{Kim \bgroup et al.\egroup
  }2004]{BioNERkim2004introduction}
Jin-Dong Kim, Tomoko Ohta, Yoshimasa Tsuruoka, Yuka Tateisi, and Nigel Collier.
\newblock 2004.
\newblock Introduction to the bio-entity recognition task at jnlpba.
\newblock In {\em Proceedings of the international joint workshop on natural
  language processing in biomedicine and its applications}, pages 70--75.
  Association for Computational Linguistics.

\bibitem[\protect\citename{Kim \bgroup et al.\egroup }2016]{NYU2016character}
Yoon Kim, Yacine Jernite, David Sontag, and Alexander~M Rush.
\newblock 2016.
\newblock Character-aware neural language models.
\newblock In {\em AAAI}, pages 2741--2749.

\bibitem[\protect\citename{Knox \bgroup et al.\egroup }2010]{Drugbank}
Craig Knox, Vivian Law, Timothy Jewison, Philip Liu, Son Ly, Alex Frolkis,
  Allison Pon, Kelly Banco, Christine Mak, Vanessa Neveu, et~al.
\newblock 2010.
\newblock Drugbank 3.0: a comprehensive resource for ‘omics’ research on
  drugs.
\newblock {\em Nucleic acids research}, 39(suppl\_1):D1035--D1041.

\bibitem[\protect\citename{Krallinger \bgroup et al.\egroup
  }2015]{CHEMdnerkrallinger2015}
Martin Krallinger, Obdulia Rabal, Florian Leitner, Miguel Vazquez, David
  Salgado, Zhiyong Lu, Robert Leaman, Yanan Lu, Donghong Ji, Daniel~M Lowe,
  et~al.
\newblock 2015.
\newblock The chemdner corpus of chemicals and drugs and its annotation
  principles.
\newblock {\em Journal of cheminformatics}, 7(S1):S2.

\bibitem[\protect\citename{Kuru \bgroup et al.\egroup
  }2016]{Character_multilingual2016}
Onur Kuru, Ozan~Arkan Can, and Deniz Yuret.
\newblock 2016.
\newblock Charner: Character-level named entity recognition.
\newblock In {\em Proceedings of COLING 2016, the 26th International Conference
  on Computational Linguistics: Technical Papers}, pages 911--921.

\bibitem[\protect\citename{Lample \bgroup et al.\egroup
  }2016]{lample2016neural}
Guillaume Lample, Miguel Ballesteros, Sandeep Subramanian, Kazuya Kawakami, and
  Chris Dyer.
\newblock 2016.
\newblock Neural architectures for named entity recognition.
\newblock {\em arXiv preprint arXiv:1603.01360}.

\bibitem[\protect\citename{Leaman and Gonzalez}2008]{leaman2008banner}
Robert Leaman and Graciela Gonzalez.
\newblock 2008.
\newblock Banner: an executable survey of advances in biomedical named entity
  recognition.
\newblock In {\em Biocomputing 2008}, pages 652--663. World Scientific.

\bibitem[\protect\citename{Lei \bgroup et al.\egroup
  }2013]{Chinese_clinicalNER}
Jianbo Lei, Buzhou Tang, Xueqin Lu, Kaihua Gao, Min Jiang, and Hua Xu.
\newblock 2013.
\newblock A comprehensive study of named entity recognition in chinese clinical
  text.
\newblock {\em Journal of the American Medical Informatics Association},
  21(5):808--814.

\bibitem[\protect\citename{Li \bgroup et al.\egroup }2005]{SVM_CoNLL2003}
Yaoyong Li, Kalina Bontcheva, and Hamish Cunningham.
\newblock 2005.
\newblock Svm based learning system for information extraction.
\newblock In {\em Deterministic and statistical methods in machine learning},
  pages 319--339. Springer.

\bibitem[\protect\citename{Li \bgroup et al.\egroup }2015]{li2015component}
Yanran Li, Wenjie Li, Fei Sun, and Sujian Li.
\newblock 2015.
\newblock Component-enhanced chinese character embeddings.
\newblock {\em arXiv preprint arXiv:1508.06669}.

\bibitem[\protect\citename{Limsopatham and Collier}2016]{Character_twitter_NER}
Nut Limsopatham and Nigel~Henry Collier.
\newblock 2016.
\newblock Bidirectional lstm for named entity recognition in twitter messages.

\bibitem[\protect\citename{Ling \bgroup et al.\egroup }2015]{CHAR_POS_SOA2015}
Wang Ling, Tiago Lu{\'\i}s, Lu{\'\i}s Marujo, Ram{\'o}n~Fernandez Astudillo,
  Silvio Amir, Chris Dyer, Alan~W Black, and Isabel Trancoso.
\newblock 2015.
\newblock Finding function in form: Compositional character models for open
  vocabulary word representation.
\newblock {\em arXiv preprint arXiv:1508.02096}.

\bibitem[\protect\citename{Liu \bgroup et al.\egroup }2015]{liu2015effects}
Shengyu Liu, Buzhou Tang, Qingcai Chen, and Xiaolong Wang.
\newblock 2015.
\newblock Effects of semantic features on machine learning-based drug name
  recognition systems: word embeddings vs. manually constructed dictionaries.
\newblock {\em Information}, 6(4):848--865.

\bibitem[\protect\citename{Luo}2015]{Luo_microsoft}
2015.
\newblock {\em Joint Named Entity Recognition and Disambiguation}, September.

\bibitem[\protect\citename{Ma and Hovy}2016]{ma2016end}
Xuezhe Ma and Eduard Hovy.
\newblock 2016.
\newblock End-to-end sequence labeling via bi-directional lstm-cnns-crf.
\newblock {\em arXiv preprint arXiv:1603.01354}.

\bibitem[\protect\citename{Malouf}2002]{CoNLL2002markov}
Robert Malouf.
\newblock 2002.
\newblock Markov models for language-independent named entity recognition,
  proceedings of the 6th conference on natural language learning.
\newblock {\em August}, 31:1--4.

\bibitem[\protect\citename{Marcus \bgroup et al.\egroup }1993]{PTB_English}
Mitchell~P Marcus, Mary~Ann Marcinkiewicz, and Beatrice Santorini.
\newblock 1993.
\newblock Building a large annotated corpus of english: The penn treebank.
\newblock {\em Computational linguistics}, 19(2):313--330.

\bibitem[\protect\citename{Mikolov \bgroup et al.\egroup
  }2013]{WordEmb_mikolov2013}
Tomas Mikolov, Kai Chen, Greg Corrado, and Jeffrey Dean.
\newblock 2013.
\newblock Efficient estimation of word representations in vector space.
\newblock {\em arXiv preprint arXiv:1301.3781}.

\bibitem[\protect\citename{Misawa \bgroup et al.\egroup
  }2017]{misawa2017character}
Shotaro Misawa, Motoki Taniguchi, Yasuhide Miura, and Tomoko Ohkuma.
\newblock 2017.
\newblock Character-based bidirectional lstm-crf with words and characters for
  japanese named entity recognition.
\newblock In {\em Proceedings of the First Workshop on Subword and Character
  Level Models in NLP}, pages 97--102.

\bibitem[\protect\citename{Nadeau and Sekine}2007]{nadeau2007survey}
David Nadeau and Satoshi Sekine.
\newblock 2007.
\newblock A survey of named entity recognition and classification.
\newblock {\em Lingvisticae Investigationes}, 30(1):3--26.

\bibitem[\protect\citename{Nadeau \bgroup et al.\egroup
  }2006]{nadeau2006unsupervised}
David Nadeau, Peter~D Turney, and Stan Matwin.
\newblock 2006.
\newblock Unsupervised named-entity recognition: Generating gazetteers and
  resolving ambiguity.
\newblock In {\em Conference of the Canadian Society for Computational Studies
  of Intelligence}, pages 266--277. Springer.

\bibitem[\protect\citename{Nguyen \bgroup et al.\egroup
  }2016]{nguyen2016vietnamese}
TS~Nguyen, LM~Nguyen, and XC~Tran.
\newblock 2016.
\newblock Vietnamese named entity recognition at vlsp 2016 evaluation campaign.
\newblock In {\em Proceedings of The Fourth International Workshop on
  Vietnamese Language and Speech Processing}.

\bibitem[\protect\citename{Ohta \bgroup et al.\egroup }2002]{ohta2002genia}
Tomoko Ohta, Yuka Tateisi, and Jin-Dong Kim.
\newblock 2002.
\newblock The genia corpus: An annotated research abstract corpus in molecular
  biology domain.
\newblock In {\em Proceedings of the second international conference on Human
  Language Technology Research}, pages 82--86. Morgan Kaufmann Publishers Inc.

\bibitem[\protect\citename{Passos \bgroup et al.\egroup
  }2014]{McLLum_CRF_2014lexicon}
Alexandre Passos, Vineet Kumar, and Andrew McCallum.
\newblock 2014.
\newblock Lexicon infused phrase embeddings for named entity resolution.
\newblock {\em arXiv preprint arXiv:1404.5367}.

\bibitem[\protect\citename{Patil \bgroup et al.\egroup
  }2016]{Indian_survey_NER}
Nita Patil, Ajay~S Patil, and BV~Pawar.
\newblock 2016.
\newblock Survey of named entity recognition systems with respect to indian and
  foreign languages.
\newblock {\em International Journal of Computer Applications}, 134(16).

\bibitem[\protect\citename{Pham and Le-Hong}2017]{pham2017end}
Thai-Hoang Pham and Phuong Le-Hong.
\newblock 2017.
\newblock End-to-end recurrent neural network models for vietnamese named
  entity recognition: Word-level vs. character-level.
\newblock {\em arXiv preprint arXiv:1705.04044}.

\bibitem[\protect\citename{Piskorski \bgroup et al.\egroup
  }2017]{multilingual2017first}
Jakub Piskorski, Lidia Pivovarova, Jan {\v{S}}najder, Josef Steinberger, Roman
  Yangarber, et~al.
\newblock 2017.
\newblock The first cross-lingual challenge on recognition, normalization and
  matching of named entities in slavic languages.
\newblock In {\em Proceedings of the 6th Workshop on Balto-Slavic Natural
  Language Processing}. Association for Computational Linguistics.

\bibitem[\protect\citename{Plank \bgroup et al.\egroup
  }2016]{Goldberg2016multilingual}
Barbara Plank, Anders S{\o}gaard, and Yoav Goldberg.
\newblock 2016.
\newblock Multilingual part-of-speech tagging with bidirectional long
  short-term memory models and auxiliary loss.
\newblock {\em arXiv preprint arXiv:1604.05529}.

\bibitem[\protect\citename{Pradhan \bgroup et al.\egroup }2013]{ONTO_notes}
Sameer Pradhan, Alessandro Moschitti, Nianwen Xue, Hwee~Tou Ng, Anders
  Bj{\"o}rkelund, Olga Uryupina, Yuchen Zhang, and Zhi Zhong.
\newblock 2013.
\newblock Towards robust linguistic analysis using ontonotes.
\newblock In {\em Proceedings of the Seventeenth Conference on Computational
  Natural Language Learning}, pages 143--152.

\bibitem[\protect\citename{Rabiner and Juang}1986]{HMM1986introduction}
Lawrence Rabiner and B~Juang.
\newblock 1986.
\newblock An introduction to hidden markov models.
\newblock {\em ieee assp magazine}, 3(1):4--16.

\bibitem[\protect\citename{Rajeev~Sangal and Singh}2008]{IndianlangNER}
Dipti Misra~Sharma Rajeev~Sangal and Anil~Kumar Singh, editors.
\newblock 2008.
\newblock {\em Proceedings of the IJCNLP-08 Workshop on Named Entity
  Recognition for South and South East Asian Languages}.
\newblock Asian Federation of Natural Language Processing, Hyderabad, India,
  January.

\bibitem[\protect\citename{Rockt{\"a}schel \bgroup et al.\egroup
  }2013]{rocktaschel2013wbi}
Tim Rockt{\"a}schel, Torsten Huber, Michael Weidlich, and Ulf Leser.
\newblock 2013.
\newblock Wbi-ner: The impact of domain-specific features on the performance of
  identifying and classifying mentions of drugs.
\newblock In {\em SemEval@ NAACL-HLT}, pages 356--363.

\bibitem[\protect\citename{Saha \bgroup et al.\egroup }2009]{SAHA2009905}
Sujan~Kumar Saha, Sudeshna Sarkar, and Pabitra Mitra.
\newblock 2009.
\newblock Feature selection techniques for maximum entropy based biomedical
  named entity recognition.
\newblock {\em Journal of Biomedical Informatics}, 42(5):905 -- 911.
\newblock Biomedical Natural Language Processing.

\bibitem[\protect\citename{Santos and Cardoso}2007]{santos2007Portugese}
Diana Santos and Nuno Cardoso.
\newblock 2007.
\newblock Reconhecimento de entidades mencionadas em portugu{\^e}s:
  Documenta{\c{c}}{\~a}o e actas do harem, a primeira avalia{\c{c}}{\~a}o
  conjunta na {\'a}rea.

\bibitem[\protect\citename{Santos and Guimaraes}2015]{santos2015boosting}
Cicero Nogueira~dos Santos and Victor Guimaraes.
\newblock 2015.
\newblock Boosting named entity recognition with neural character embeddings.
\newblock {\em arXiv preprint arXiv:1505.05008}.

\bibitem[\protect\citename{Schapire}2013]{adaBoost2013explaining}
Robert~E Schapire.
\newblock 2013.
\newblock Explaining adaboost.
\newblock In {\em Empirical inference}, pages 37--52. Springer.

\bibitem[\protect\citename{Segura~Bedmar \bgroup et al.\egroup
  }2013]{segura2013semeval}
Isabel Segura~Bedmar, Paloma Mart{\'\i}nez, and Mar{\'\i}a Herrero~Zazo.
\newblock 2013.
\newblock Semeval-2013 task 9: Extraction of drug-drug interactions from
  biomedical texts (ddiextraction 2013).
\newblock Association for Computational Linguistics.

\bibitem[\protect\citename{Shaalan}2014]{Arabic2014survey}
Khaled Shaalan.
\newblock 2014.
\newblock A survey of arabic named entity recognition and classification.
\newblock {\em Computational Linguistics}, 40(2):469--510.

\bibitem[\protect\citename{Sharnagat}2014]{MLNERsurvey}
Rahul Sharnagat.
\newblock 2014.
\newblock Named entity recognition: A literature survey.
\newblock {\em Center For Indian Language Technology}.

\bibitem[\protect\citename{Strassel \bgroup et al.\egroup
  }2003]{strassel2003multilingual}
Stephanie Strassel, Alexis Mitchell, and Shudong Huang.
\newblock 2003.
\newblock Multilingual resources for entity extraction.
\newblock In {\em Proceedings of the ACL 2003 workshop on Multilingual and
  mixed-language named entity recognition-Volume 15}, pages 49--56. Association
  for Computational Linguistics.

\bibitem[\protect\citename{Takeuchi and Collier}2002]{SVM2002_MUC6}
Koichi Takeuchi and Nigel Collier.
\newblock 2002.
\newblock Use of support vector machines in extended named entity recognition.
\newblock In {\em proceedings of the 6th conference on Natural language
  learning-Volume 20}, pages 1--7. Association for Computational Linguistics.

\bibitem[\protect\citename{Tjong Kim~Sang and
  De~Meulder}2003]{tjong2003introduction}
Erik~F Tjong Kim~Sang and Fien De~Meulder.
\newblock 2003.
\newblock Introduction to the conll-2003 shared task: Language-independent
  named entity recognition.
\newblock In {\em Proceedings of the seventh conference on Natural language
  learning at HLT-NAACL 2003-Volume 4}, pages 142--147. Association for
  Computational Linguistics.

\bibitem[\protect\citename{Tjong Kim~Sang}2002]{sang2002introduction}
Erik~F Tjong Kim~Sang.
\newblock 2002.
\newblock Introduction to the conll-2002 shared task: language-independent
  named entity recognition, proceedings of the 6th conference on natural
  language learning.
\newblock {\em August}, 31:1--4.

\bibitem[\protect\citename{Uzuner \bgroup et al.\egroup
  }2007]{uzuner2007evaluating}
{\"O}zlem Uzuner, Yuan Luo, and Peter Szolovits.
\newblock 2007.
\newblock Evaluating the state-of-the-art in automatic de-identification.
\newblock {\em Journal of the American Medical Informatics Association},
  14(5):550--563.

\bibitem[\protect\citename{Uzuner \bgroup et al.\egroup }2011]{I2B2_license}
{\"O}zlem Uzuner, Brett~R South, Shuying Shen, and Scott~L DuVall.
\newblock 2011.
\newblock 2010 i2b2/va challenge on concepts, assertions, and relations in
  clinical text.
\newblock {\em Journal of the American Medical Informatics Association},
  18(5):552--556.

\bibitem[\protect\citename{Xu \bgroup et al.\egroup }2017]{xu2017bidirectional}
Kai Xu, Zhanfan Zhou, Tianyong Hao, and Wenyin Liu.
\newblock 2017.
\newblock A bidirectional lstm and conditional random fields approach to
  medical named entity recognition.
\newblock In {\em International Conference on Advanced Intelligent Systems and
  Informatics}, pages 355--365. Springer.

\bibitem[\protect\citename{Yadav \bgroup et al.\egroup }2018]{yadav2018deep}
Vikas Yadav, Rebecca Sharp, and Steven Bethard.
\newblock 2018.
\newblock Deep affix features improve neural named entity recognizers.
\newblock In {\em Proceedings of the Seventh Joint Conference on Lexical and
  Computational Semantics}, pages 167--172.

\bibitem[\protect\citename{Yan \bgroup et al.\egroup
  }2016]{yan2016multilingual}
Shao Yan, Christian Hardmeier, and Joakim Nivre.
\newblock 2016.
\newblock Multilingual named entity recognition using hybrid neural networks.
\newblock In {\em The Sixth Swedish Language Technology Conference (SLTC)}.

\bibitem[\protect\citename{Yang \bgroup et al.\egroup }2016]{yang2016multi}
Zhilin Yang, Ruslan Salakhutdinov, and William Cohen.
\newblock 2016.
\newblock Multi-task cross-lingual sequence tagging from scratch.
\newblock {\em arXiv preprint arXiv:1603.06270}.

\bibitem[\protect\citename{Yang \bgroup et al.\egroup }2017]{yang2017transfer}
Zhilin Yang, Ruslan Salakhutdinov, and William~W Cohen.
\newblock 2017.
\newblock Transfer learning for sequence tagging with hierarchical recurrent
  networks.
\newblock {\em arXiv preprint arXiv:1703.06345}.

\bibitem[\protect\citename{Yin \bgroup et al.\egroup }2016]{yin2016multi}
Rongchao Yin, Quan Wang, Peng Li, Rui Li, and Bin Wang.
\newblock 2016.
\newblock Multi-granularity chinese word embedding.
\newblock In {\em Proceedings of the 2016 Conference on Empirical Methods in
  Natural Language Processing}, pages 981--986.

\bibitem[\protect\citename{Zhang and Elhadad}2013]{zhangBIOmedunsupervised}
Shaodian Zhang and No{\'e}mie Elhadad.
\newblock 2013.
\newblock Unsupervised biomedical named entity recognition: Experiments with
  clinical and biological texts.
\newblock {\em Journal of biomedical informatics}, 46(6):1088--1098.

\bibitem[\protect\citename{Zhou and Su}2002]{HMMzhou2002named}
GuoDong Zhou and Jian Su.
\newblock 2002.
\newblock Named entity recognition using an hmm-based chunk tagger.
\newblock In {\em proceedings of the 40th Annual Meeting on Association for
  Computational Linguistics}, pages 473--480. Association for Computational
  Linguistics.

\end{thebibliography}

\end{document}